%% file: main.tex
\documentclass[sigconf]{acmart}
\usepackage[linesnumbered,ruled,vlined]{algorithm2e}
\usepackage{xcolor}
\usepackage{colortbl}
\usepackage{adjustbox}   
\usepackage{booktabs}    
\usepackage{multirow}    
\usepackage{subcaption}
\AtBeginDocument{%
  }

\setcopyright{acmlicensed}
\copyrightyear{2026}
\acmYear{2026}
\setcopyright{cc}
\setcctype{by-nc-nd}
\acmConference[MM '26] {Proceedings of the 34th ACM International Conference on Multimedia}{November 10--14, 2026}{Rio de Janeiro, Brazil}
\acmBooktitle{Proceedings of the 34th ACM International Conference on Multimedia (MM '26), November 10--14, 2026, Rio de Janeiro, Brazil}
\acmDOI{10.1145/3767308.3835256}
\acmISBN{979-8-4007-2213-4/2026/11}

\acmSubmissionID{2022}
\begin{document}

\title{ZoomV: Temporal Zoom-in for Efficient Long Video Understanding}


\author{Yuan Zhang}
\authornote{Equal contribution.}
\affiliation{%
  \institution{School of Computer Science,\\ Peking University}
  \city{Beijing}
  \country{China}
}
\email{zhangyuan@alumni.pku.edu.cn}

\author{Junwen Pan}
\authornotemark[1]
\affiliation{%
  \institution{ByteDance Inc.}
  \city{Beijing}
  \country{China}}
\email{panjunwen@bytedance.com}

\author{Rui Zhang}
\authornotemark[1]
\affiliation{%
  \institution{ByteDance Inc.}
  \city{Beijing}
  \country{China}}
\email{zhangrui.1994@bytedance.com}

\author{Xin Wan}
\affiliation{%
  \institution{ByteDance Inc.}
  \city{Beijing}
  \country{China}}
\email{wanxinhh@gmail.com}

\author{Qizhe Zhang}
\affiliation{%
  \institution{School of Computer Science,\\ Peking University}
  \city{Beijing}
  \country{China}
}
\email{theia@pku.edu.cn}

\author{Ming Lu}
\affiliation{%
  \institution{School of Computer Science,\\ Peking University}
  \city{Beijing}
  \country{China}
}
\email{lu199192@gmail.com}

\author{Qi She}
\authornote{Correspondence to: Qi She and Shanghang Zhang.}
\affiliation{%
  \institution{ByteDance Inc.}
  \city{Beijing}
  \country{China}}
\email{sheqi.roger@bytedance.com}

\author{Shanghang Zhang}
\authornotemark[2]
\affiliation{%
  \institution{School of Computer Science,\\ Peking University}
  \city{Beijing}
  \country{China}
}
\email{shanghang@pku.edu.cn}

\renewcommand{\shortauthors}{Yuan Zhang et al.}

\input{sec/0_abstract}


\begin{teaserfigure}
  \centering
  \Description{Illustration of human-like interaction for long-video understanding.}
  \includegraphics[width=0.95\textwidth]{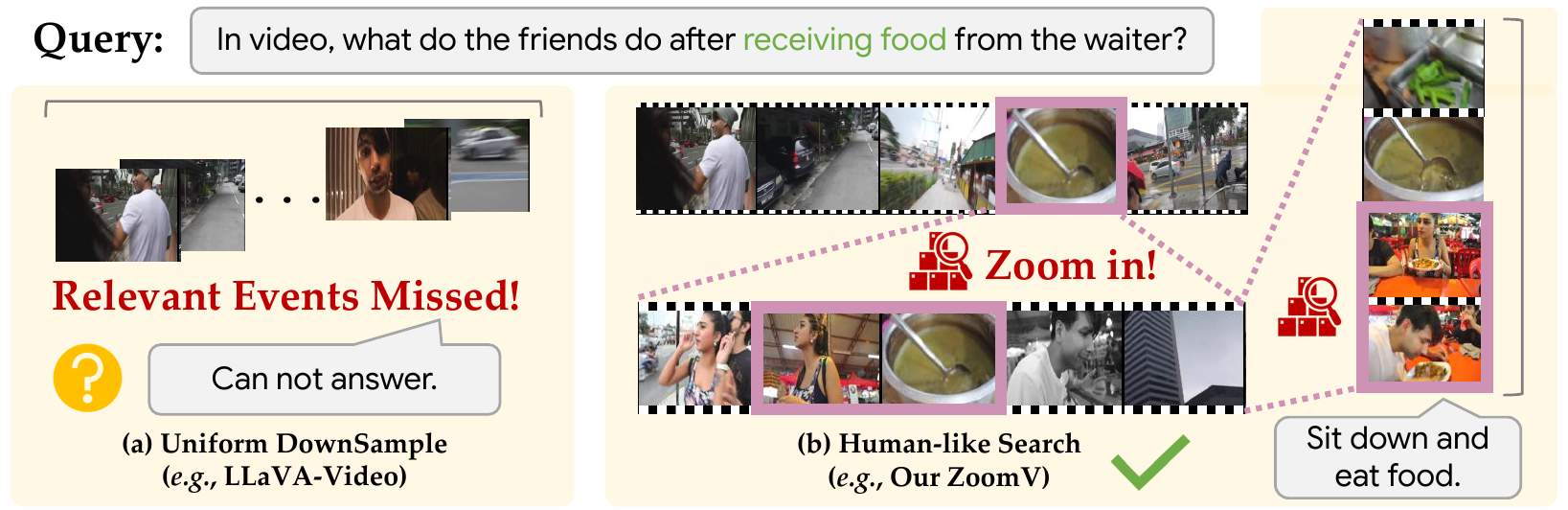}
  \vspace{-2mm}
  \caption{\textbf{Illustration of human-like interaction for long-video understanding.} It divides hour-long videos into manageable sub-events and searches within query-aware segments.}
  \label{fig:teaser}
\end{teaserfigure}


\maketitle

\input{sec/1_introduction}

\input{sec/2_related_work}

\input{sec/3_method}

\input{sec/4_experiment}

\input{sec/5_conclusion}

\begin{acks}
This work was supported by the National Natural Science Foundation of China (62476011), the Beijing Natural Science Foundation (L252060), and the Beijing Major Science and Technology Project (Z191100010618003).
\end{acks}

\bibliographystyle{ACM-Reference-Format}
\bibliography{main}



\end{document}

%% file: sec/0_abstract.tex
\begin{abstract}
Long video understanding poses a fundamental challenge for large video-language models (LVLMs) due to the overwhelming number of frames and the risk of losing essential context through naive downsampling.
Inspired by the way humans watch videos on mobile phones, \textit{constantly zooming in on frames of interest}, we propose \textbf{ZoomV}, a query-aware temporal zoom-in framework designed for efficient and accurate long video understanding. 
Specifically, ZoomV operates in three stages: (1) Temporal interests grounding: guided by the query, ZoomV retrieves relevant events and their associated temporal windows as candidates. (2) Event interests spotlighting: within pools of candidate windows, each window is scored through the model itself reflection and filtered accordingly, where higher-confidence windows are more representative. (3) Compact representation: the selected events are encoded and temporally downsampled to preserve critical semantics while significantly reducing redundancy.
Extensive experiments demonstrate that ZoomV substantially outperforms prior \textbf{video-agent–style} approaches. On temporal grounding, ZoomV unlocks the latent capability of LVLMs, achieving an $\textbf{11.8\%}$ \textbf{mIoU} gain on Charades-STA. Remarkably, ZoomV further boosts accuracy on LVBench by $\textbf{9.7\%}$, underscoring its effectiveness on long-video benchmarks.
\end{abstract}

%% file: sec/1_introduction.tex
\section{Introduction}
\label{sec:intro}
Long-form video understanding, involving frame sequences that can span minutes to hours, presents a fundamental challenge in computer vision.
While large video-language models (LVLMs) have shown impressive performance on video-language tasks \cite{zhang2024llavaVideo, mvbench, activitynet, activityqa, qvhighlights, wu2024longvideobench}, they still struggle with long videos. On the one hand, feeding all frames into the model results in a rapid increase in computational cost and memory usage, making naive full-frame processing infeasible. On the other hand, aggressive temporal downsampling risks discarding critical context, often resulting in severe visual hallucinations. For instance, the advanced LLaVA-Video uniformly samples only $64$ frames regardless of video duration, leading to a significant loss of detailed temporal information, especially in hour-long videos \cite{zhang2024llavaVideo}. Therefore, it is crucial to identify \textit{a sufficient number of the most relevant frames in a prompt-aware manner} for reliable long-video understanding.

Existing attempts to scale LVLMs to long videos can be mainly grouped into two directions. The first is token sparsification, reducing sequence length by discarding a subset of visual tokens or patches \cite{llama-vid, zhang2024sparsevlm, ma2025mmg, zhang2025beyond, zhang2025chainv}. While this improves computational efficiency, it inevitably compromises the holistic integrity of frames, and \textit{often leads to noticeable performance degradation}. The second is video-agent approaches, which typically assemble \textit{a pipeline of heterogeneous models}, \emph{e.g.}, \cite{wang2024videoagent, wang2025videotree} using EVA-CLIP \cite{sun2023eva} for retrieval, BLIP \cite{li2022blip} for captioning, and GPT-4 \cite{achiam2023gpt} for reasoning. Although such modular systems alleviate sequence-length constraints, their reliance on disparate models makes them inefficient, resource-intensive, and non-end-to-end.

To address the limitations of both paradigms, we take inspiration from human cognitive strategies \cite{sweller1994cognitive, zhang2026freekd+, zhang2023avatar, zacks2007event, vstar, shen2024zoomeye}, particularly the way humans selectively zoom in on relevant visual content, and introduce our method, dubbed ZoomV, a query-aware temporal zoom-in framework for efficient long-video understanding. 
As illustrated in Figure \ref{fig:teaser} (b), humans review videos broadly to find relevant clues, then gradually focus on more specific sub-events for detailed inspection. Importantly, when the necessary information is not immediately clear, humans may revisit multiple candidate sub-events iteratively, grounding and validating their relevance until they find satisfactory answers. 
Therefore, ZoomV imitates the behavior and progressively divides the video timeline into coarse-grained events and finer-grained sub-events, enabling efficient human-like search. Specifically, ZoomV unfolds in three progressive stages. 

In the first stage, to identify subtle temporal details within promising sub-events accurately, we first need to ground temporal interest. Previous LVLMs \cite{videochat, zhang2024llavaVideo} have incorporated temporal instructions to improve understanding, yet they do not effectively align visual and temporal cues. In contrast, our ZoomV employs a TemporaLink that explicitly embeds temporal information into visual frame representations, enabling LVLMs to precisely associate visual content with corresponding timestamps. Additionally, to alleviate quantization errors introduced by frame sampling, we optimize the absolute timestamp representation, stabilizing temporal learning and enhancing grounding performance. The model retrieves query-relevant events and their associated temporal windows as candidate regions along the video timeline, providing a coarse yet comprehensive coverage of potentially relevant content.

In the second stage, event interests spotlighting, each candidate window is evaluated through the model’s self-reflection mechanism dubbed TemporaLight, which assigns confidence scores to spotlight the most representative windows while filtering out less relevant ones. As humans hierarchically search through time, they continuously reflect on whether a specific sub-event warrants deeper inspection. Similarly, we leverage the self-reflection capability of LVLMs to guide the search process. Recent studies \cite{LinHE22Teaching, kadavath2022language, zheng2023judging, Zhang24:GenRM} have demonstrated that LLMs effectively assess their prediction confidence through additional multiple-choice or yes/no questions. Inspired by these findings, we first identify that LVLMs inherently possess a similar self-reflection capability—``\textit{they know what they do not know}.'' This selective scoring process ensures that only high-confidence sub-events proceed to the next stage.

\begin{figure*}[t]
\centering
\Description{An illustrative view of ZoomV.}
\includegraphics[width=.95\textwidth]{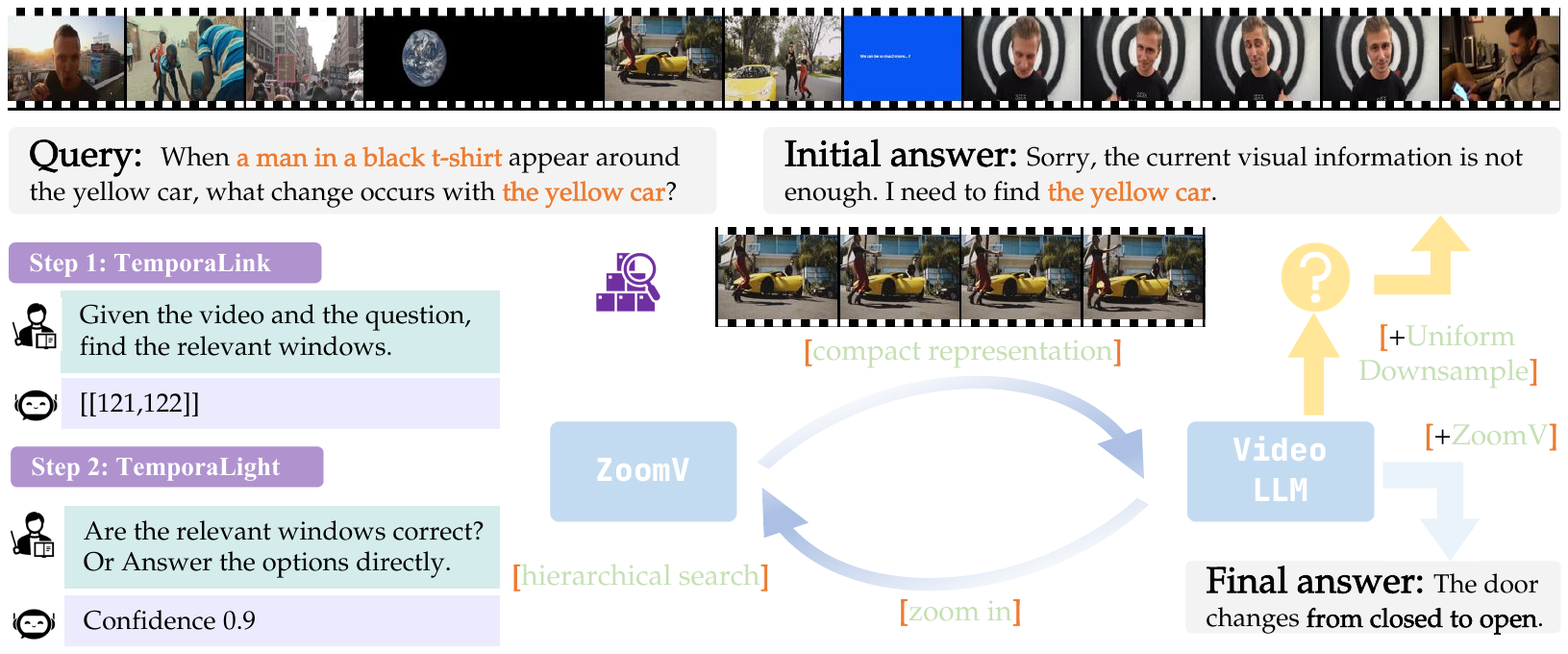}
\vspace{-2mm}
\caption{\textbf{Overview of ZoomV.} It enables Video LLMs to understand long videos more efficiently and accurately.}
\vspace{-2mm}
\label{fig:input_output_steps}
\end{figure*}

In the third stage, compact representation, the spotlighted events are encoded and temporally downsampled into condensed representations that preserve critical semantics while substantially reducing redundancy. Together, these stages enable ZoomV to progressively zoom in \textbf{from coarse-grained temporal coverage to fine-grained} and compact representations, achieving efficient and accurate long-video understanding. Extensive experiments demonstrate its superior performance across various challenging benchmarks, including VideoMME \cite{videomme}, MLVU \cite{MLVU}, and LongVideoBench \cite{wu2024longvideobench}. Notably, on the highly challenging LVBench dataset with hour-long videos \cite{wang2024lvbench}, ZoomV achieves new state-of-the-art accuracy, substantially surpassing previous methods.  
In summary, our contributions are threefold:
\begin{itemize}
    \item We reveal that LVLMs inherently own strong self-reflection abilities, enabling reflection-guided prioritization of temporal search and serve as a key foundation of our framework.
    
    \item We propose \textbf{TemporaLink}, which unlocks latent temporal grounding capabilities in LVLMs. Despite its simplicity, TemporaLink achieves an $11.8\%$ mIoU gain over state-of-the-art temporal grounding models.
    
    \item We propose \textbf{ZoomV}, a query-aware hierarchical temporal zoom-in framework comprising TemporaLink and 
    \textbf{TemporaLight}, which mimics human coarse-to-fine exploration for efficient long-video understanding. ZoomV improves accuracy on LVBench from $41.8\%$ to $51.5\%$.
\end{itemize}

%% file: sec/2_related_work.tex
\section{Related Work}
\label{sec:related_work}

\subsection{Long Video Understanding}
Understanding lengthy videos for LVLMs is challenging due to the need to store and extract information from numerous frames. 
One common line involves using language as a bridge to summarize videos into concise captions \cite{VideoReCap,zhang2023simple-longrange}, resulting in the omission of vital visual signals.  
Another widely studied line involves memory-based methods for compressing video features into a limited memory bank, which is achieved by continually updating the memory bank during visual encoding \cite{moviechat}. 
The memory bank has also been applied to real-time streaming video understanding, potentially enabling an unlimited length of frames while maintaining a constant space footprint \cite{flashvstream}. 
A major drawback of these methods is their oversight of video duration and information density, particularly when utilizing a fixed space for a memory bank. 
For instance, Flash-VStream compresses both brief $10$-second clips and hour-long movies into the same $681$ tokens \cite{flashvstream}.
Besides, these black box methods lack interpretability, as it is hard to verify whether pertinent details are accurately retrieved for reasoning.
Another line of work reduces video tokens through visual compression, including spatial token selection \cite{jin2024chat,zhang2024sparsevlm,zhang2025beyond}, temporal frame selection \cite{yu2025frame,tang2025adaptive,zhang2025loss}, and joint spatio-temporal selection~\cite{LongVU}. These methods typically serve as pre-processing strategies before LLM inference. In contrast, ZoomV leverages the inherent temporal grounding and reflection ability, enabling task-aware and hierarchical selection.

\subsection{Temporal Grounding}
Temporal Grounding localizes video segments relevant to textual queries \cite{charades-sta, localizing, qvhighlights} or questions \cite{chen2024rextime,nextgqa}. Current LVLM-based approaches typically rely on heavy customization to handle continuous time, such as specialized time tokens \cite{wang2024grounded,vtgllm,timechat}, auxiliary temporal modules \cite{momentor,hawkeye}, or extensive reinforcement training (RL) \cite{li2025videochatr1, wang2025timer1, pan2025timesearch}.
In contrast, ZoomV shows that standard LVLMs inherently possess precise grounding capabilities when properly guided.
ZoomV unlocks this potential through a simple TemporaLink and coarse-to-fine zooming mechanism, achieving state-of-the-art performance without architectural modifications or sophisticated RL.

\subsection{VideoAgents}
VideoAgent is a LLM agent that understands videos by using customized structured tools \cite{wang2024videoagent, localizing, wang2025videotree}.
Previous research usually needs one more model for the agent pipeline. 
For instance, \cite{wang2024videoagent} introduces a prompt-driven video QA agent that employs extra vision-language retrieval models (\emph{e.g.}, CLIP) to ground key frames during reasoning. Building on this, VideoTree \cite{wang2025videotree} designs a hierarchical tree-style search by clustering frames with visual features, enabling structured exploration. Both approaches still rely on caption models to provide frame-level descriptions. 
In contrast, our ZoomV avoids additional captioning or grounding models and directly leverages LVLMs to predict continuous temporal windows. Moreover, the tree-like structure of ZoomV is constructed purely over simple temporal segments, without requiring extra visual feature models or clustering algorithms.
We also highlight two concurrent works with hierarchical search strategies. UniTime~\cite{universalvideo} exhaustively performs temporal grounding on all video segments, while VideoChat-R1.5~\cite{li2025videochatr1} employs a fixed three-step grounding strategy. 
In contrast, ZoomV designs an \emph{adaptive} search mechanism with reflection-guided early termination: TemporaLight assesses confidence to avoid exhaustive searches, and the priority queue enables flexible exploration with backtracking to alternative paths. These innovations allow ZoomV to achieve superior efficiency and accuracy by focusing computation on relevant temporal windows.

%% file: sec/3_method.tex
\section{Proposed Approach: ZoomV}
\label{sec:methodology}

This section presents the ZoomV framework (Figure \ref{fig:input_output_steps}), which equips LVLMs with a human-inspired, hierarchical temporal zoom-in mechanism. We first introduce \textbf{TemporaLink}, a temporally augmented representation that binds timestamps with visual frames for accurate temporal grounding. Next, we leverage the inherent self-reflection capability of LVLMs to develop \textbf{TemporaLight}, a reflection-guided hierarchical search algorithm that progressively zooms from coarse events to fine sub-events to generate compact representations for efficient long-video understanding.

\subsection{Preliminary: Autoregressive Modeling}
Our ZoomV is built upon an autoregressive LVLM backbone, which sequentially predicts tokens conditioned on visual and textual contexts. 
An autoregressive LVLM generates an output sequence $\mathbf{y} = (y_1, y_2, \dots, y_L)$ with length $L$ given a text condition $\textbf{x}$ and a video condition $\mathbf{v}$ by predicting tokens one at a time based on the previously generated tokens.
Assuming that the LVLM is parameterized by $\theta$, the conditional probability distribution of generating a sequence $\textbf{y}$ given context $\mathbf{x}$ and $\mathbf{v}$ is defined as
\begin{equation}
    p_\theta(\mathbf{y} | \mathbf{v}, \mathbf{x}) = \prod_{i=1}^{L} p_\theta(y_i | \mathbf{v}, \mathbf{x}, \mathbf{y}_{<i}), 
    \label{eq:auto_regressive}
\end{equation}
where $\mathbf{y}_{<1} = \emptyset$ and $\mathbf{y}_{<t} = (y_1, y_2, \dots, y_{t-1})$. 
Taking video question answering (VQA) as an example, an LVLM predicts the answer distribution $p_\theta(\mathbf{a} \mid \mathbf{v}, \mathbf{q}, I_q)$, 
where $\mathbf{q}$ denotes the input question and 
$I_q = \text{``\texttt{Answer the following questions related to this video}}$'' \\
\noindent serves as the instruction. Here, $\mathbf{v}$ represents a sequence of $T$ downsampled frame tokens extracted from the original video, which are transformed by a dedicated visual encoder and projector into visual tokens. In the following sections, we extend this autoregressive formulation to model both the grounding and reflection mechanisms within a unified framework.

\begin{figure}[t]
    \begin{center}
    \Description{Illustration of TemporaLink.}
    \includegraphics[width=0.46\textwidth]{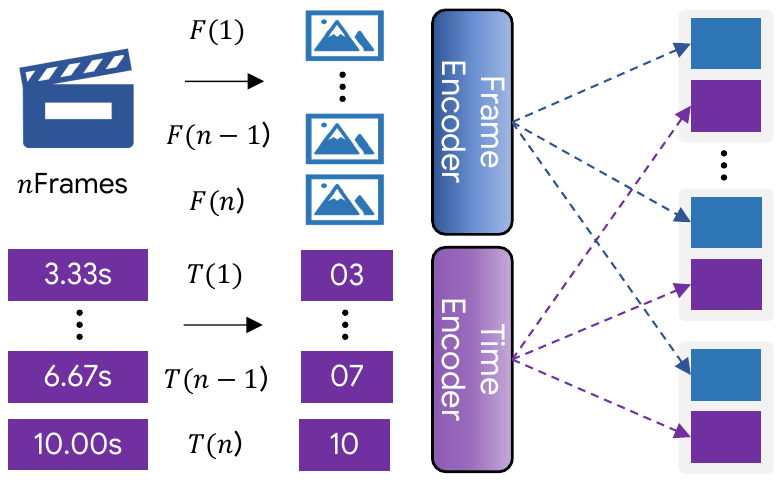}
    \end{center}
    \vspace{-2mm}
    \caption{\textbf{Illustration of TemporaLink.}}
    \label{fig:TAFR}
    \vspace{-2mm}
\end{figure}

\subsection{Interest Grounding by TemporaLink}
\label{sec:temporal_grounding}

Temporal interest grounding aims to identify the most relevant temporal windows according to the query, modeling continuous numerical timestamps as discrete digit generations \citep{timechat, vtgllm}.
The task process is defined as: (1) Given a query \(\mathbf{q}\) and the grounding instruction $I_g=$ ``\texttt{Find the relevant windows}'', the model predicts text sequence \( p_\theta(\mathbf{w} | \mathbf{v}, \mathbf{q}, I_g) \); (2) Then the text sequence $\mathbf{w}$ is turned into a set of time ranges ${W}=[(s_1, e_1), \dots, (s_K, e_K) ]$ with size $K$, where $s_k, e_k$ signifies the start and end timestamps of $k$-th target window clip.
However, LVLMs naturally struggle to accurately handle numerical tasks, especially in temporal tasks involving precise numerical comparisons \citep{numerologic, xie24:OrderMatters}. To alleviate this challenge and effectively activate the inherent temporal grounding capability of LVLMs, we propose a simple yet effective enhancement module, dubbed \textbf{TemporaLink}, which explicitly binds timestamps with visual frame representations.

Specifically, given a downsampled video represented by frames \((f_1, f_2, \dots, f_T )\) and their corresponding fractional timestamps \((t_1, t_2, \allowbreak \dots, t_T)\), \emph{e.g.}, \texttt{(0.00, 3.33, 6.67, 10.00)}, we first round these timestamps to the nearest integer. Then, to ensure a consistent token representation in TemporaLink, we apply left-zero padding, resulting in timestamps like \texttt{(00, 03, 07, 10)}:
\begin{equation}
    \Tilde{t}_i = \text{Pad} (\text{Round}(t_i)).
\end{equation}
Next, we extract frame visual features through a visual encoder $\mathcal{V}$ with a projection module \citep{zhang2024llavaVideo}. To embed the absolute timestamp into each frame feature, as illustrated in Figure \ref{fig:TAFR}, we directly concatenate these features with their corresponding absolute timestamp embeddings:
\begin{equation}
    \Tilde{\mathbf{v}}_i =  \text{concat}(\mathcal{V}(f_i), \mathcal{T}(\Tilde{t}_i)), ~ \Tilde{\mathbf{v}}_i \in \mathbb{R}^{(N+P)\times D},
\end{equation}
where $\mathcal{T}$ denotes the embedding layer of the LLM, $D$ represents the embedding dimension, and $N$, $P$ denote the number of visual frame tokens and padded timestamp tokens, respectively. Following the above design, we perform refinement training on the LVLM. 
During this process, manually annotated timestamps are further aligned with the rounded timestamps to mitigate quantization errors, as detailed in Appendix B. 
By explicitly linking visual frames with timestamps, TemporaLink not only significantly enhances the temporal grounding capability of LVLMs but also provides a solid foundation for the subsequent spotlighting and selection stages.

\subsection{Interests Spotlighting by TemporaLight}
\label{sec:temporal_reflection}

\begin{figure}[t]
    \begin{center}
        \Description{A scatter plot showing that reflection probabilities correlate positively with mIoU and classification accuracy.}
        \includegraphics[width=1\linewidth]{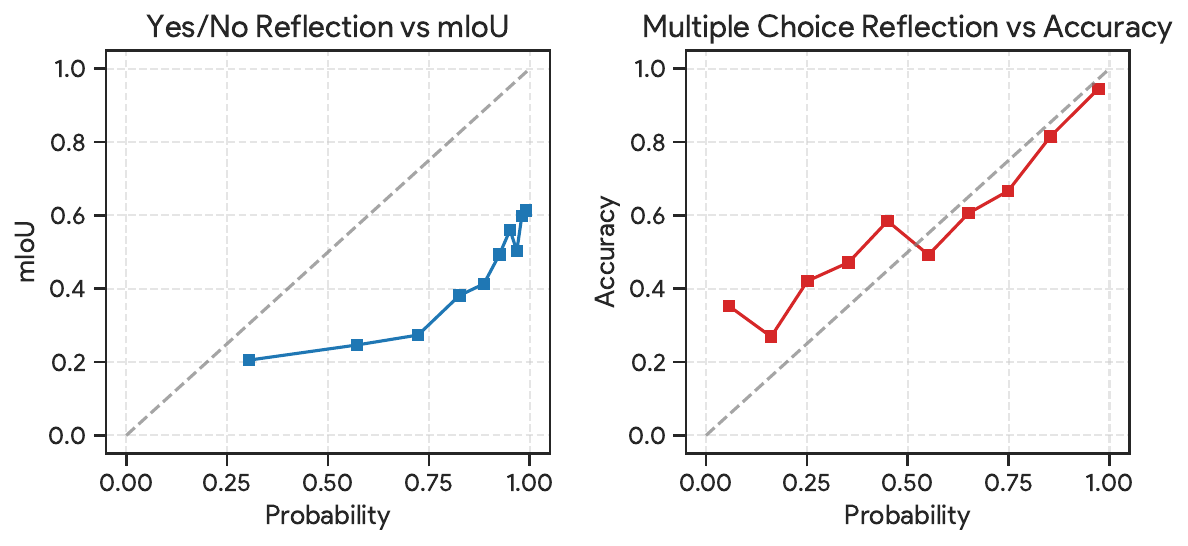}
    \end{center}
    \vspace{-2mm}
    \caption{\textbf{Reflection probabilities correlate with performance.}}
    \vspace{-3mm}
    \label{fig:self_reflection}
\end{figure}

After obtaining the candidate temporal windows, we need to validate them and highlight the most suitable ones. 
Previous research has demonstrated that generative LLMs can evaluate the correctness of their predictions through self-reflection mechanisms \citep{LinHE22Teaching, zheng2023judging, Zhang24:GenRM, zhang2024unveiling}. These models can produce well-calibrated confidence scores for Yes/No and multiple-choice questions. We extend this observation from text-based LLMs to LVLMs and propose TemporaLight to assess the validity of temporal spotlight predictions. Here, there are two forms of validation.

For Yes/No type reflection validation, given a question \(\mathbf{q}\), a TemporaLight prediction $W$, and reflection instruction $I_\text{tf}=$``\texttt{Are the proposed relevant windows correct?}'', the probability is formulated as
\begin{equation}
    c = p_\theta(\text{``\texttt{Yes}''} | \mathbf{v}, \mathbf{q}, W, I_\text{tf}).
    \label{eq:reflection}
\end{equation}
The Yes/No reflection confidence positively correlates with grounding accuracy (mIoU), thus providing an intrinsic measure of spotlight correctness without human annotations (Figure \ref{fig:self_reflection}, top).
For multiple-choice reflection validation, the reflection confidence score is defined by selecting the maximum prediction probability from multiple choices. Given a set of candidate answers, the reflection confidence is computed as:
\begin{equation}
    c = \text{max}~ \left\{p_\theta( o | \mathbf{v}, \mathbf{q}, W, I_\text{mc}) \right\}, { o \in \text{(``\texttt{A}'', ``\texttt{B}'', \dots)} },
    \label{eq:reflection2}
\end{equation}
where $I_\text{mc}$ is the reflection instruction, \emph{e.g.}, ``\texttt{Answer the options directly}’’. The calibration analysis in Figure \ref{fig:self_reflection} (bottom) further confirms that LVLMs produce reliable reflection scores, especially at high-confidence levels. 
In summary, the above two modes indicate that LVLM inherently knows whether ``relevant windows can be found'' and ``whether questions can be correctly answered.''

\input{sec/algorithm}

Notably, ZoomV iteratively performs temporal interest grounding to identify query-aware temporal events. At each step, the TemporaLight module is employed to evaluate the confidence $c$ of the currently candidate windows. If the reflection confidence $c$ is below a predefined threshold $\epsilon$, we hierarchically split the event into three equal-sized overlapping sub-events (``\texttt{beginning}'', ``\texttt{middle}'', and ``\texttt{end}''.), recursively exploring sub-events prioritized by reflection confidence scores $c$, as illustrated in Algorithm \ref{alg:guided_search}.
In this algorithm, ZoomV adopts a priority queue $\mathbf{PQ}$ to organize the order of sub-event searches, which allows backtracking to coarser-grained events to explore alternative search paths when current sub-events still do not yield enough information.
This hierarchical search terminates either when the confidence exceeds a threshold hyper-parameter $\epsilon$, or when the sub-event duration falls below a minimal threshold $\Delta$. The highlighted temporal windows with the highest reflection confidence are utilized for the following video understanding tasks.

\subsection{Event Compact Representation}
Receiving the spotlighted events, ZoomV constructs compact representations for them, where high-confidence windows are encoded through the LVLM’s visual encoder to extract visual embeddings, followed by temporal downsampling to retain key frames and discard less informative ones. The embeddings are then aggregated into event-level representations that maintain temporal order and salient visual-textual correlations. By concentrating computation on the most representative sub-events, ZoomV achieves effective long-video understanding without excessive memory or latency costs, substantially contributing to its overall performance.

%% file: sec/algorithm.tex
\begin{algorithm}[t]
\small
\SetAlgoLined
\SetAlgoNoEnd
\SetKw{Init}{Initialize:}
\SetKw{Import}{Import:}
\SetKwProg{Fn}{def}{:}{}

\KwIn{
$\mathbf{v}$, $\mathbf{q}$, \\
$\Delta$ is the sub-event duration threshold, 
$\epsilon$ is the confidence threshold. 
}

\Init{
    \begin{itemize}
        \item $\mathbf{PQ}$: a priority queue prioritised by confidence.
        \item $W$: the candidate optimal window; $c$: best confidence.
        \item $W, c \leftarrow \textsc{SpotlightReflect}(\mathbf{v})$
        \item $\textsc{Enqueue}(\mathbf{PQ}, \mathbf{v}, W, \text{priority}=c)$
    \end{itemize}
}
\Fn{$\textsc{SpotlightReflect}(\mathbf{v}_i$)}{
        ${W}_i=\textsc{Ground}(\textsc{FrameSample}(\mathbf{v}_i), \mathbf{q}, I_g)$ \;
        \If{question $\mathbf{q}$ is open-ended}{
            ${c}_i=\textsc{Reflect}(\mathbf{v}, \mathbf{q}, W, I_\text{tf})$ \tcp*[h]{Yes/No}\;
        }
        \Else {
            ${c}_i = \textsc{Reflect}(\mathbf{v}, \mathbf{q}, W, I_\text{mc})$ \tcp*[h]{MC}\;
        }
        \Return{${W}_i, {c}_i$}
}
\While{$\mathbf{PQ}$ is not empty} {
    \tcp*[h]{Pop sub-event with top priority}\;
    $\mathbf{v}_i, {W}_i, {c}_i \leftarrow \textsc{Dequeue}(\mathbf{PQ})$ \;
    \If{$ {c}_i \ge {c} $}{
      $c \leftarrow {c}_i$ \; ~ $W \leftarrow {W}_i$ \;
    }
   \If{$ {c}_i \ge \epsilon$}{
      break \tcp*[h]{stop criterion}\;
   }
    \For{$\mathbf{v}_j \in \{\texttt{begin, mid, end} \}$  of $\mathbf{v}_i$}{
        \If{$\textsc{Length}(\mathbf{v}_j)\geq \Delta$}{
            $W_j, c_j \leftarrow \textsc{SpotlightReflect}(\mathbf{v}_j)$\;
            $\textsc{Enqueue}(\mathbf{PQ}, \mathbf{v}_j, W_j, priority=c_j)$
        }
    }
}
\KwOut{$W$, the optimal temporal windows}
\caption{TemporaLight Hierarchical Search}
\label{alg:guided_search}
\end{algorithm}

%% file: sec/4_experiment.tex
\section{Experiments}
\label{sec:experiments}

\subsection{Experiment Settings}
\paragraph{(1) Benchmarks.}
We evaluate ZoomV across three tasks spanning \textbf{seven} subtasks. 
For \textbf{Video QA}, we validate ZoomV on long-video multiple-choice benchmarks (\emph{e.g.,} MLVU~\cite{MLVU}, LongVideoBench~\cite{wu2024longvideobench}, and LVBench~\cite{wang2024lvbench}) ranging from minutes to hours, as well as the short-video VideoMME~\cite{videomme}. 
For \textbf{Temporal Sentence Grounding}, we perform zero-shot evaluation on Charades-STA~\cite{charades-sta} and ActivityNet-Captions~\cite{activitynet}. 
For \textbf{Temporal Question Grounding}, we test on ReXTime~\cite{chen2024rextime} to assess temporal and causal reasoning.

\paragraph{(2) Models.} 
We apply our ZoomV to the LLaVA-Video~\cite{zhang2024llavaVideo}, InternVL2.5~\cite{chen2024internvl}, and Qwen2.5-VL~\cite{qwen2.5-VL}, three different model architectures for generality. The training of TemporaLink is both completed within eight hours using $128$ A$100$ GPUs. To enhance TemporaLight capabilities without sacrificing general performance, we apply LoRA~\cite{lora} with a rank of $32$ to the LLM for fine-tuning.

\paragraph{(3) Video Input Format.} Within the identified time window \( W \), interest frames are densely sampled from the spotlighted segments for video understanding. These dense frames are \textbf{appended} after the globally sparsely sampled frames, thereby retaining the ability to answer questions about the global video context. In our experiments, the number of global frames is set to $64$, while the maximum number of spotlight frames provided by ZoomV is $16$.

\input{tab/main}

\subsection{Main Results}

\paragraph{(1) Video Question Answering Tasks.}
As illustrated in Table~\ref{tab:long_video_benchmarks}, ZoomV consistently boosts existing open-source LVLMs across a wide range of benchmarks. On short-video tasks such as MLVU, ZoomV maintains competitive performance, ensuring that its long-video enhancements do not come at the cost of short-duration understanding. The advantage becomes \textit{more pronounced on long-video datasets}: for instance, ZoomV improves InternVL2.5 on LVBench (average duration $4101$ seconds) from $41.8\%$ to $51.5\%$, and \textbf{makes it surpass all prior methods}. Similar gains are observed on LongVideoBench (+$2.7\%$ accuracy) and VideoMME-Long (+$1.7\%$), highlighting its effectiveness in handling videos lasting up to several hours. Furthermore, the consistent improvements across different LVLM backbones, such as an $\textbf{8.7}\%$ and $\textbf{11.3}\%$ boost on LVBench with LLaVA-Video and Qwen2.5-VL, demonstrate the robustness and versatility of our ZoomV as a general solution for temporal search in long video understanding.

\paragraph{(2) Temporal Grounding Tasks.} 
\input{tab/main_2}

As shown in Table~\ref{tab:temporal_grounding_sota}, ZoomV delivers substantial improvements over existing grounding-oriented LVLMs. On standard benchmarks such as Charades-STA and ActivityNet Captions, it achieves an average mIoU gain of $\textbf{11.8}\%$ compared with the previous state-of-the-art (\emph{e.g.}, GroundedVideo-LLM). Beyond these datasets, ZoomV demonstrates even stronger advantages on ReXTime, where it boosts mIoU by $\textbf{8.6}\%$, Recall@$0.5$ by $\textbf{9.6}\%$, and nearly \textbf{doubles} VQA accuracy from $40.0\%$ to $76.5\%$ relative to TimeChat, clearly highlighting its ability to reason over complex temporal event structures.

\begin{figure*}[t]
    \centering
    \begin{subfigure}[t]{0.49\textwidth}
        \centering
        \Description{Effectiveness of TemporaLink.}
        \includegraphics[width=\textwidth]{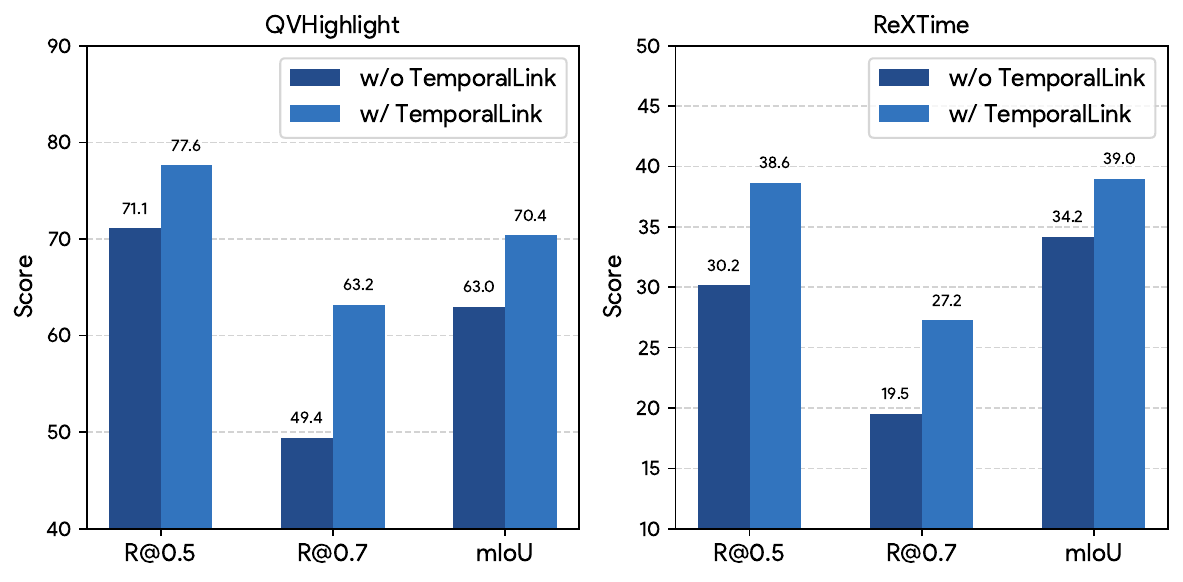}
        \caption{\textbf{Effectiveness of TemporaLink.}}
        \label{fig:ablation_TemporaLink}
    \end{subfigure}
    \hfill
    \begin{subfigure}[t]{0.49\textwidth}
        \centering
        \Description{Robustness to video length via TemporaLink.}
        \includegraphics[width=\textwidth]{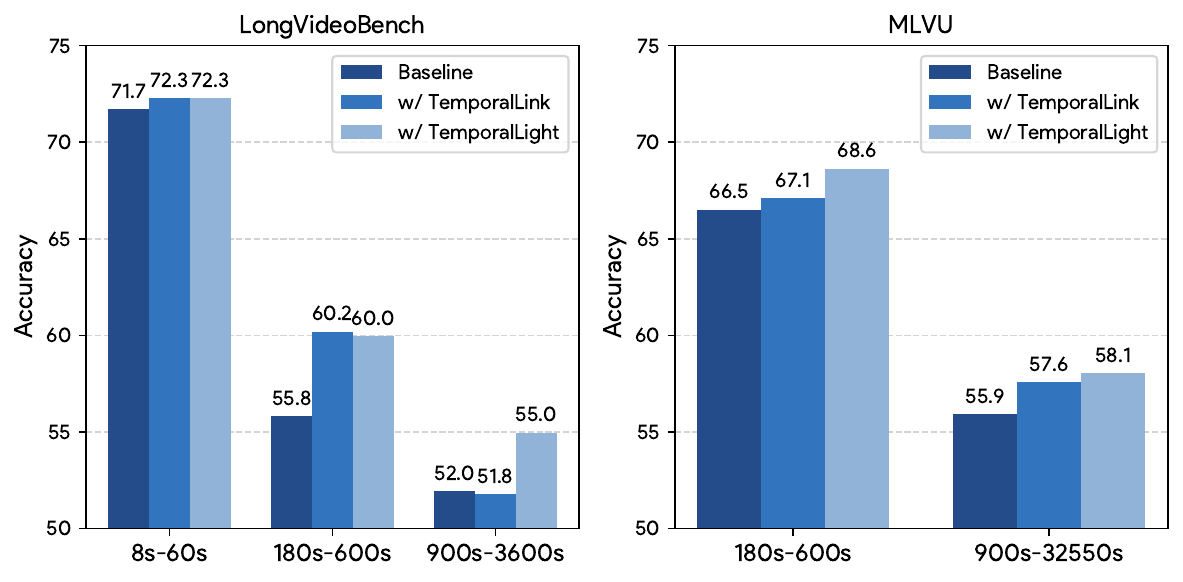}
        \caption{\textbf{Robustness to video length via TemporaLink.}}
        \label{fig:ablation_longvideo_qa}
    \end{subfigure}
    \vspace{-2mm}
    \caption{\textbf{Ablation studies on temporal grounding and ultra-long video length.}}
    \vspace{-2mm}
    \label{fig:ablation_combined}
\end{figure*}

\begin{figure*}[t]
    \centering
    \Description{Impact of the spotlight frames on LongVideoBench.}
    \includegraphics[width=1\textwidth]{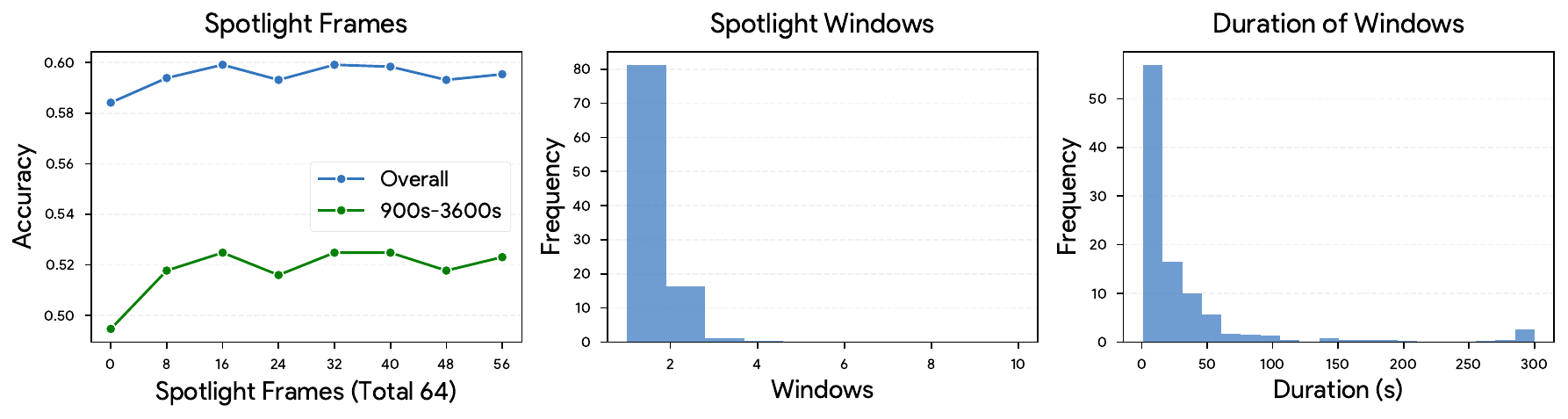}
    \vspace{-2mm}
    \caption{
        \textbf{Impact of the spotlight frames on LongVideoBench.}
    }
    \vspace{-2mm}
    \label{fig:abl_frame_nums}
\end{figure*}

\section{Analysis}
\label{sec:ablation_study}

\subsection{Effectiveness of TemporaLink}
We validate TemporaLink by comparing it to the widely used Time Instructions on various datasets. As shown in Figure \ref{fig:ablation_TemporaLink}, compared to the time instruction, TemporaLink significantly improves the ability of moment retrieval, especially at high IoU thresholds. Specifically, when replacing TemporaLink with time instruction on the QVHighlight, R@$0.7$ drops sharply by $13.8\%$ while R@$0.5$ drops $6.5\%$. Although previous LVLMs are capable of recognizing relevant events, they struggle to accurately establish associations between events and timelines without TemporaLink. Furthermore, TemporaLink provides consistent improvements on the challenging ReXTime, which requires a strong ability to reason across time.

\subsection{Effectiveness of TemporaLight} To validate our TemporaLight effectiveness, we employ the LLaVA-Video model as the baseline, and equip it with our two types of reflections to watch the difference. The results are reported in Table \ref{fig:ab_TemporaLight}. We find that our TemporaLight enhances the baseline by $8.8\%$ on LVBench, and $2.7\%$ on LongVideoBench. Additionally, multiple-choice reflection, which offers more options for the model, shows better performance on video understanding tasks.

\begin{table*}[t]
    \centering
    \caption{\textbf{Detailed Comparison of ZoomV on efficiency and accuracy on EgoSchema.}}
    \vspace{-2mm}
    \setlength{\tabcolsep}{10pt}
    \renewcommand{\arraystretch}{1.}
    \begin{tabular}{l|ccccc|cc}
    \toprule
    \textbf{Method} & \textbf{grounding (s)} & \textbf{reflect (s)} & \textbf{caption (s)} & \textbf{keyfr. (s)} & \textbf{QA (s)} & \textbf{overall (s)} & \textbf{acc. (\%)} \\
    \midrule
    VideoTree \cite{wang2025videotree} & -- & -- & 1.6 & 4.4 & 1.8 & 7.8 & 63.6 \\
    ZoomV & 1.6 & 1.9 & -- & -- & 1.9 & 5.4 & 63.7 \\
    \bottomrule
    \end{tabular}
    \label{tab:com_eff}
\end{table*}

\begin{table}[t]
\centering
\setlength{\tabcolsep}{1.2pt}
\renewcommand{\arraystretch}{1}
\caption{\textbf{Analysis of TemporaLight on video understanding.}}
\begin{tabular}{l|ccc}
\toprule
Method & LVBench & LongVideo & LongVideo-Long \\ 
 \midrule
Base (LLaVA-Video) & 41.3 & 58.3 & 48.4 \\ 
Yes/No Reflection & 43.3 & 61.0 & 50.4 \\
Multi-choice Reflection & 50.1 & 61.0 & 51.8 \\ 
\bottomrule
\end{tabular}
\vspace{-2mm}
\label{fig:ab_TemporaLight}
\end{table}

\subsection{Ablation Study of Frames Zoomed in}
We conduct experiments by varying the number of frames we zoom in, while keeping the frame budget fixed at $64$.
As shown in Figure \ref{fig:abl_frame_nums} (left), introducing spotlight frames yields a significant boost in accuracy for both general cases and long videos. 
Our results suggest that $16$ is an optimal setting, as it preserves global awareness while ensuring precise event retrieval.
For a more in-depth understanding, we further analyze the number of spotlight windows and their duration distributions in Figure \ref{fig:abl_frame_nums} (middle and right). 
The histogram of spotlight window counts reveals that most examples require only one or two spotlight windows, suggesting that many questions can be effectively answered with a small number of targeted events. 
Moreover, the spotlight duration histogram indicates that a majority of spotlighted events are relatively short (under 50 seconds).
These findings highlight that a small number of well-chosen short spotlights is sufficient for significant improvements in long-video understanding, validating the effectiveness of our reflection-guided temporal search strategy in selecting relevant video moments efficiently.

\subsection{Robustness to Ultra-Long Video Lengths} ZoomV shows noticeable improvements for video understanding models in Figure \ref{fig:ablation_longvideo_qa}. Specifically, for medium-length videos (\textit{i.e.}, \texttt{180s-600s}), simply applying the TemporaLink to supplement event details can yield consistent gains. Empirically, despite the significant loss of temporal dynamics in frame sampling, TemporaLink can still identify windows relevant to the questions based on limited visual cues. As the video length increases (\textit{i.e.}, over \texttt{900s} ), it becomes increasingly challenging for TemporaLink to focus on useful events through sparse frames, and our search strategies are needed and result in significant improvements. Notably, for short videos, the framework maintains original performance as expected.

\subsection{Efficiency of ZoomV}
While we boost LVLM performance via ZoomV, we uphold efficiency optimizations to ensure practicality. Initially, training a high-quality ZoomV on the LLaVA-Video model within $80$ epochs only requires $8$ hours utilizing $128$ NVIDIA A$100$ GPUs. Furthermore, ZoomV introduces minimal additional latency during inference, and the runtime per search step is $3483$ms. We further optimize the multi-turn search with the prefix cache, as shown in Appendix Table 2. Eventually, we compare our method with other video–agent–style approaches (\emph{e.g.}, VideoTree). As shown in Table \ref{tab:com_eff}, while achieving higher accuracy, our method requires only $5.4$s under the optimal parameters on a typical long-video dataset, compared to $7.8$s for VideoTree, yielding an acceleration of approximately $30.8\%$.

\begin{figure}[t]
    \vspace{0mm}
    \begin{center}
        \Description{Runtime comparison across different video durations.}
        \includegraphics[width=0.92\linewidth]{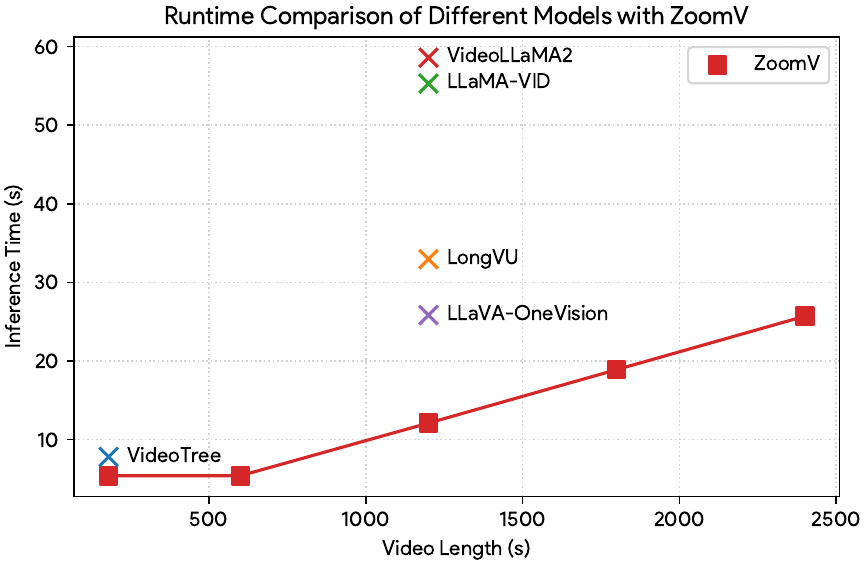}
    \end{center}
    \vspace{-2mm}
    \caption{\textbf{Runtime comparison across different video durations.} We report the worst-case scenario, demonstrating superior efficiency compared to fixed-cost long-video models.}
    \vspace{-2mm}
    \label{fig:plot_cost_inference_time}
\end{figure}

To further validate the efficiency, we compare the inference time of ZoomV with other agents and long-video understanding models across varying durations. As illustrated in Figure \ref{fig:plot_cost_inference_time}, we report the runtime of ZoomV in the worst-case scenario (\textit{i.e.}, searching to the finest granularity), benchmarking it against state-of-the-art models including VideoTree~\cite{wang2025videotree}, LongVU~\cite{LongVU}, LLaVA-OneVision~\cite{li2024llavaonevision}, VideoLLaMA2~\cite{video-llama}, and LLaMA-VID~\cite{llama-vid}.
ZoomV exhibits a linear scaling of inference time with video duration, yet remains highly efficient. Specifically, for short videos, ZoomV completes inference in just $\sim 5$ seconds, outperforming VideoTree. Even as the duration extends to $1200$ seconds, ZoomV requires only $\sim 12.1$ seconds. In sharp contrast, LongVU takes $33$ seconds, while VideoLLaMA2 and LLaMA-VID incur significantly higher latency. These results underscore ZoomV's superior scalability and efficiency for long-video understanding.

%% file: tab/main.tex
\definecolor{ForestGreen}{RGB}{34,139,34}
\newcommand{\fg}[1]{\mathbf{\mathcolor{ForestGreen}{#1}}}
\begin{table*}[t]
    \centering
    \caption{\textbf{Comparison of ZoomV with other LVLMs on video understanding.} The results on various short and long video benchmarks with video durations ranging from seconds to hours.}
    \vspace{-3mm}
    \setlength{\tabcolsep}{6pt}
    \renewcommand{\arraystretch}{0.95}
    \begin{tabular}{lc c |c cc cc}
    \toprule  
    \multirow{2}[2]{*}{\textbf{Model}} & \multirow{2}[2]{*}{\textbf{Size}}& \multirow{2}[2]{*}{\textbf{\#Frames}} & \multirow{2}[2]{*}{\textbf{MLVU}} & \multirow{2}[2]{*}{\textbf{LongVideoBench}} & \multicolumn{2}{c}{\textbf{VideoMME}} & \multirow{2}[2]{*}{\textbf{LVBench}} \\ 
    \cmidrule(lr){6-7}
    & & & & & Long & Overall \\
    \midrule
    \textbf{Average Duration} & & & 651$s$ & 473$s$ &  {2386$s$} & {1010$s$}& 4101$s$ \\
    \midrule
    \rowcolor{gray!10} \multicolumn{8}{c}{\textit{Proprietary LVLMs}} \\
    GPT-4V \citep{GPT4V} &-&-& 49.2 & 60.7  & 53.5 & 59.9 & - \\
    GPT-4o \citep{GPT4o} &-&-& 64.6 & 66.7& 65.3 & 71.9 & 34.7 \\
    Gemini-1.5-Pro \citep{team2024gemini} &-&-& 61.8 & 64.4 & 67.4 & 75.0 & 33.1 \\
    \midrule
    \rowcolor{gray!10} \multicolumn{8}{c}{\textit{Open-Sourced LVLMs}} \\
    InternVL2 \citep{chen2024internvl} & 8B & - & 64.0 &  54.6 &  - & - & - \\
    Qwen2-VL \citep{Qwen2VL} & 7B & - & - & - &  - & 63.3 & - \\
    Qwen2.5-VL \citep{bai2025qwen2} & 7B & 768 & - & - & - & 65.1 & 45.3 \\
    LLaVA-OneVision \citep{li2024llavaonevision} & 7B & - &  64.7 & 56.3 &  - & 58.2 & - \\
    LLaVA-OneVision \citep{li2024llavaonevision} & 72B & - &  68.0 & 61.3 &  - & - & 26.9 \\
    \midrule
    \rowcolor{gray!10} \multicolumn{8}{c}{\textit{Long-Video LVLMs}} \\ 
    VideoLLaMA2 \citep{video-llama} & 7B & 72 & 48.5&- &42.1&47.9 & - \\
    LongVA \citep{zhang2024long} & 7B & 128 & 56.3 &- & 46.2&52.6 & - \\
    LLaMA-VID \citep{llama-vid} & 7B & 1 FPS & 33.2 &  - & - & - & 23.9 \\
    Oryx \citep{liu2024oryx} & 7B & 1 FPS & 67.5 & 55.3 & 50.3 & 58.3 & - \\
    Oryx-1.5 \citep{liu2024oryx} & 7B & 1 FPS & 67.5 & 56.3 & 51.2 & 58.8 & - \\
    LongVU \citep{LongVU} & 7B & 1 FPS & 65.4 & 59.5 & 52.4 & 60.6 & - \\
    \midrule
    \rowcolor{gray!10} \multicolumn{8}{c}{\textit{Video Agents}} \\
    VideoAgent (GPT-4) \citep{wang2024videoagent} & - & 87 & - &-  & 49.0 & 56.0 & - \\
    VideoTree (GPT-4o) \citep{wang2025videotree} & - & 98 & - &-  & 53.1 & - & - \\
    UniTime \citep{universalvideo} & 7B & 128 & 66.5 & 56.5 & - & - & - \\
    LLaVA-Video \citep{zhang2024llavaVideo} & 7B & 80 & 64.4  & 58.3 &  52.4 & 63.4 & 41.3 \\
    \multicolumn{2}{l}{ { \textit{w}/ \textbf{ZoomV}}} & 64+16 &  68.1 $\fg{(\uparrow 3.7)}$ &  60.9 $\fg{(\uparrow 2.6)}$ &  53.9 $\fg{(\uparrow 1.5)}$ & 64.0 $\fg{(\uparrow 0.6)}$ & 50.0  $\fg{(\uparrow 8.7)}$  \\
    InternVL2.5 \citep{chen2024internvl} & 8B & 80 & 67.1 & 60.6 & 52.2 & 63 & 41.8 \\
    \multicolumn{2}{l}{ { \textit{w}/ \textbf{ZoomV}}} & 64+16 & 70.0 $\fg{(\uparrow 2.9)}$ & 63.3 $\fg{(\uparrow 2.7)}$ & 53.9 $\fg{(\uparrow 1.7)}$ & 64.4 $\fg{(\uparrow 1.4)}$ & 51.5 $\fg{(\uparrow 9.7)}$ \\
    Qwen2.5-VL \citep{qwen2.5-VL} & 7B & 80 & 65.9 & 59.0 & 52.0 & 63.5 & 40.0 \\
    \multicolumn{2}{l}{ { \textit{w}/ \textbf{ZoomV}}} & 64+16 & 67.0 $\fg{(\uparrow 1.1)}$ & 61.0 $\fg{(\uparrow 2.0)}$ & 53.6 $\fg{(\uparrow 1.6)}$ & 63.6 $\fg{(\uparrow 0.1)}$ & 51.3 $\fg{(\uparrow 11.3)}$ \\
    InternVL3 & 8B & 80 & 67.5 & 60.8 & 54.6 & 65.3 & 43.1 \\
    \multicolumn{2}{l}{\textit{w}/ \textbf{ZoomV}} & 64+16 & 68.5 $\fg{(\uparrow 1.0)}$ & 63.6 $\fg{(\uparrow 2.8)}$ & 54.9 $\fg{(\uparrow 0.3)}$ & 65.6 $\fg{(\uparrow 0.3)}$ & 51.6 $\fg{(\uparrow 8.5)}$ \\
    \bottomrule
    \end{tabular}
    \vspace{-2mm}
    \label{tab:long_video_benchmarks}
\end{table*}

%% file: tab/main_2.tex
\begin{table*}[t]
    \centering
    \caption{\textbf{Comparison of ZoomV with other LVLMs on video grounding results.} The results include two temporal-sentence and one temporal-question grounding benchmarks.}
    \vspace{-2mm}
    \setlength{\tabcolsep}{5pt}
    \renewcommand{\arraystretch}{0.95}
    \begin{tabular}{lcccccccccccc}
    \toprule
    \multirow{2}{*}{\textbf{Model}} & \multicolumn{4}{c}{\textbf{Charades-STA}} & \multicolumn{4}{c}{\textbf{ActivityNet-Captions}} & \multicolumn{4}{c}{\textbf{ReXTime}} \\
    \cmidrule(lr){2-5} \cmidrule(lr){6-9} \cmidrule(lr){10-13}
     & {R@0.3} & {R@0.5} & {R@0.7} & {mIoU} & {R@0.3} & {R@0.5} & {R@0.7} & {mIoU} & {R@0.3} & {R@0.5} &  {mIoU} & {VQA}  \\
    \midrule
    CG-DETR~\citep{moon2023correlation} &70.4 & 58.4 & 36.3 & \cellcolor{gray!10}{50.1} & - & - & - & - & 31.3 & 16.6 & \cellcolor{gray!10}{23.8} & -  \\
    UniVTG~\citep{lin2023univtg} &72.6 &60.2 & 38.6& \cellcolor{gray!10}{52.1} & - & - & - & - & 41.3 & 26.8 & \cellcolor{gray!10}{28.1} & -  \\
        \midrule
    LITA~\citep{lita} &-&-&-&-&-&-&-&-& 29.49 & 16.29 & \cellcolor{gray!10}{21.49} & \cellcolor{gray!10}{34.44} \\
    SeViLA~\citep{sevila} & 27.0 & 15.0 & 5.8 & \cellcolor{gray!10}{18.3} & 31.6 & 19.0 & 10.1 & \cellcolor{gray!10}{23.0} & - & - & - & -   \\
    {Valley~\citep{valley}} & 28.4 & 1.8 & 0.3 & \cellcolor{gray!10}{21.4} & 30.6 & 13.7 & 8.1 & \cellcolor{gray!10}{21.9} & - & - & - & - \\
    {VideoChat2~\citep{mvbench}} & 38.0 & 14.3 & 3.8 & \cellcolor{gray!10}{24.6} & 40.8 & 27.8 & 9.3 & \cellcolor{gray!10}{27.9} & - & - & - & -\\
    Momenter~\citep{momentor} & 42.6 & 26.6 & 11.6 & \cellcolor{gray!10}{28.5} & 42.9 & 23.0 & 12.4 & \cellcolor{gray!10}{29.3} & - & - & - & - \\
    {VTimeLLM~\citep{vtimellm}} & 51.0 & 27.5 & 11.4 & \cellcolor{gray!10}{31.2} & 44.0 & 27.8 & 14.3 & \cellcolor{gray!10}{30.4} & 28.8 & 17.4 & \cellcolor{gray!10}{20.1} & \cellcolor{gray!10}{36.1}\\
    TimeChat~\citep{timechat} &  {46.7} & {32.2} &  {15.7} & - & - & - & - & - & 14.4 & 7.6 & \cellcolor{gray!10}{11.6} & \cellcolor{gray!10}{40.0} \\
    HawkEye~\citep{hawkeye} & 50.6 & 31.4 & 14.5 & \cellcolor{gray!10}{{33.7}} & {49.1} & {29.3} & 10.7 & \cellcolor{gray!10}{{32.7}} & - & - & - & - \\
    GroundedVideo-LLM~\citep{wang2024grounded} & 54.2 & 36.4 & 19.7 & \cellcolor{gray!10}{36.8} & 46.2 & 30.3 & 19.0 & \cellcolor{gray!10}{36.1} & - & - & - & - \\
    \midrule
    Qwen2.5-VL~\citep{bai2025qwen2} & - & 24.2 & 11.1 & 29.0 & - & 15.8 & 7.5 & 21.1 & - & - & - & - \\
    VideoChat-R1~\citep{li2025videochatr1} & - & 70.6 & 47.2 & 59.9 & - & 33.3 & 16.7 & 35.5 & - & - & - & - \\
    Time-R1~\citep{wang2025timer1} & 78.1 & 60.8 & 35.3 & - & 58.6 & 39.0 & 21.4 & - & 31.81 & 18.46 & 22.48 & 72.1 \\
    UniTime~\citep{universalvideo} & - & 59.1 & 31.9 & 52.2 & - & 22.8 & 14.1 & 27.3 & - & - & - & - \\
    \midrule
    \textbf{Our ZoomV} & {73.6} & {52.4} & {24.5} & \cellcolor{gray!10}{{48.6}} & {61.0} & {43.0} & {26.1} & \cellcolor{gray!10}{{43.9}} & {48.4} & {36.4}  &  \cellcolor{gray!10}{36.7}  & \cellcolor{gray!10}{76.5}\\
    \bottomrule 
    \end{tabular}
    \vspace{-0.5mm}
  \label{tab:temporal_grounding_sota}
\end{table*}

%% file: sec/5_conclusion.tex
\section{Conclusion}
This paper introduces ZoomV, a novel framework for long-video understanding that emulates a human-like hierarchical temporal search. ZoomV proposes TemporaLink to retrieve key events and TemporaLight to verify predictions and guide the search direction. ZoomV achieves state-of-the-art performance across diverse video benchmarks, demonstrating significant gains in long-video QA and temporal grounding tasks. Furthermore, comprehensive ablation studies confirm the effectiveness of each component and underscore the importance of specialized designs for ultra-long video analysis. Finally, ZoomV bridges the gap between human cognitive strategies and model-based video analysis, providing a robust and interpretable solution for long video tasks.